\documentclass[runningheads]{llncs}
\usepackage{graphicx}

\usepackage{tikz}
\usepackage{comment}
\usepackage{amsmath,amssymb} 
\usepackage{color}

\usepackage[breaklinks=true,
            bookmarks=false,
            colorlinks,
            linkcolor=red,       
            anchorcolor=blue,  
            citecolor=green, 
            ]{hyperref}

\usepackage[accsupp]{axessibility}  

\usepackage[width=122mm,left=12mm,paperwidth=146mm,height=193mm,top=12mm,paperheight=217mm]{geometry}

\begin{document}
\pagestyle{headings}
\mainmatter
\def\ECCVSubNumber{2243}  

\title{NeRFocus: Neural Radiance Field for 3D Synthetic Defocus} 

\titlerunning{NeRFocus}
%
\author{Yinhuai Wang\inst{1} \and
Shuzhou Yang\inst{1} \and
Yujie Hu\inst{1} \and
Jian Zhang\inst{1,2}}
\authorrunning{W. Yinhuai et al.}
%
\institute{Peking University, Shenzhen Graduate School\inst{1}\\
Peng Cheng Laboratory\inst{2}
}
\maketitle

\begin{abstract}
Neural radiance fields (NeRF) bring a new wave for 3D interactive experiences. However, as an important part of the immersive experiences, the defocus effects have not been fully explored within NeRF. Some recent NeRF-based methods generate 3D defocus effects in a post-process fashion by utilizing multiplane technology. Still, they are either time-consuming or memory-consuming. This paper proposes a novel thin-lens-imaging-based NeRF framework that can directly render various 3D defocus effects, dubbed NeRFocus. Unlike the pinhole, the thin lens refracts rays of a scene point, so its imaging on the sensor plane is scattered as a circle of confusion (CoC). A direct solution sampling enough rays to approximate this process is computationally expensive. Instead, we propose to inverse the thin lens imaging to explicitly model the beam path for each point on the sensor plane and generalize this paradigm to the beam path of each pixel, then use the frustum-based volume rendering to render each pixel's beam path. We further design an efficient probabilistic training (p-training) strategy to simplify the training process vastly. Extensive experiments demonstrate that our NeRFocus can achieve various 3D defocus effects with adjustable camera pose, focus distance, and aperture size. Existing NeRF can be regarded as our special case by setting aperture size as zero to render large depth-of-field images. Despite such merits, NeRFocus does not sacrifice NeRF's original performance (e.g., training and inference time, parameter consumption, rendering quality), which implies its great potential for broader application and further improvement. Code and video are available at https://github.com/wyhuai/NeRFocus.
\end{abstract}

\section{Introduction}

A small camera aperture took an image usually has a very large depth-of-field (DOF), making all the objects in the scene clear and in focus. A shallow DOF is obtained if a large camera aperture is used to capture the same scene. In this condition, objects near the focal plane appear clear, while those far from the focal plane are optically blurred. This phenomenon is known as defocus effects and is prevalent in film-making and portrait photography. 

Defocus effects are depth-dependent, and if we strictly follow the principle of lens imaging~\cite{cook1984distributed}, \cite{kolb1995realistic}, the computation will be expensive. Previous image-based methods for rendering defocus usually simplify the lens imaging with the depth-dependent blur operations on discretized depth layers.
Different from these methods, NeRF~\cite{nerf} renders images by implicitly modeling the scene representation, thus naturally containing depth information for defocus effects. However, a fundamental barrier for NeRF to render defocus effects is that its imaging is based on a pinhole model, by which the whole scene is in sharp focus. To solve this limitation, we propose NeRFocus, a novel NeRF-based framework that approximates the imaging of a thin lens model. 

\begin{figure}[t]
\centering
\includegraphics[width=\linewidth]{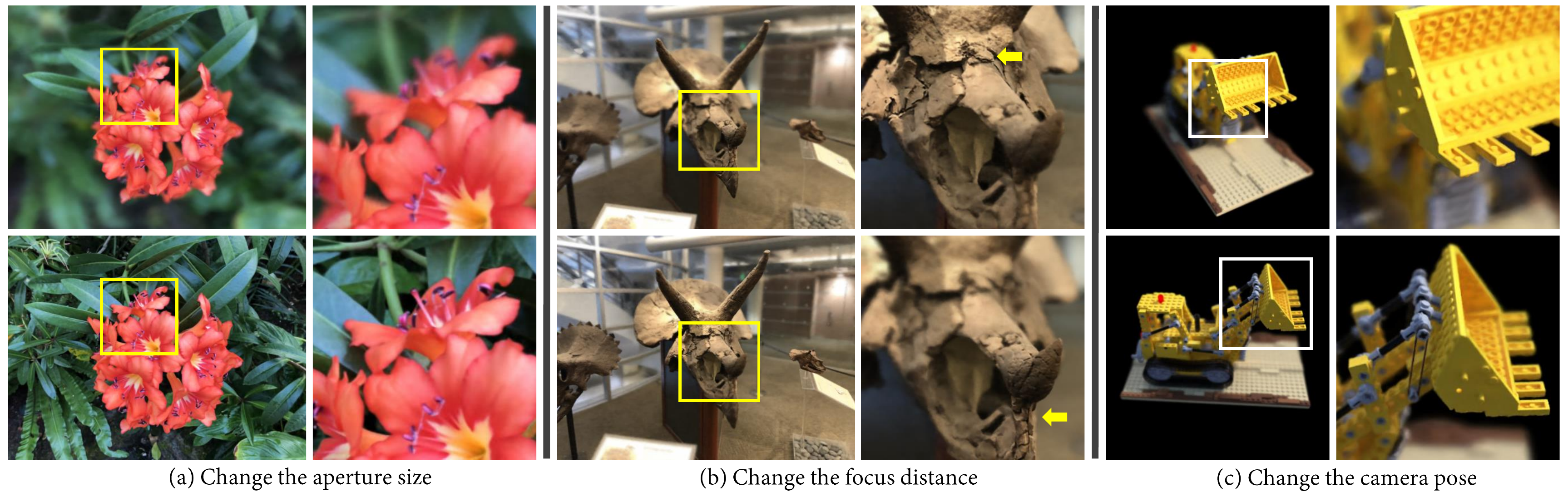}
\caption{We present a novel framework that extends a neural radiance field (NeRF) to directly render various 3D defocus effects without post-process. 
(a) We can change the aperture size to raise or reduce defocus effects. (b) Or, change the focus distance to move the depth-of-field (DOF) forward or backward, as the yellow arrows denote. (c) Our framework fully inherits Nerf's novel view synthesis and can synthesize high-quality 3D defocus effects with disentangled control on aperture, focus distance, and camera pose. \textbf{See the supplemented video for demonstrations}.}
\label{fig:teaser}
\end{figure}

In a thin lens model, a scene point scatters as a circle-of-confusion (CoC) on the sensor plane. The CoC diameter can be derived from the basic lens equation. We inverse the thin lens imaging to form the beam path for each point on the sensor plane as their beam path and generalize this paradigm to the beam path of each pixel. Specifically, we incorporate the merit of mip-NeRF~\cite{mipnerf} that extend a circle rather than a point as the receptive field of a pixel. We calculate the beam path for each point inside the circle following the inverse thin lens imaging. These beam paths finally overlap as a composite cone representing the beam path of a pixel.

To render the composite cone using volume rendering~\cite{kajiya1984ray}, \cite{max1995optical}, we divide it as frustums and assume the ``importance" distribution inside each frustum to be a 3D Gaussian, by which we can calculate the expectation of radiance and density inside each frustum. To simplify the computation, we make a critical assumption that this expectation can be directly predicted by a multilayer perceptron (MLP) given the expected value of positional encoding (PE) as input. Hence, the critical problem is, how can we make sure that the MLP can predict correctly? 

Our insight is to convert this problem into an image-based supervised training task by simply setting the aperture size as zero. In this condition, the composite cone will degrade to a cone (i.e., thin lens imaging will degrade to pinhole imaging). The rendered color of a pixel should be consistent with the ground truth (GT) pixel color. If we scale up the pixel's receptive circle, the rendering result should be equivalent to the color of GT Pixel diffused by the same scale. This diffusion can be simply implemented using a Gaussian blur kernel. Further, if we introduce different scales for training, then different scales of each frustum will be involved, which constitutes a simple solution to supervise the MLP for correct prediction.

Consequently, we propose an efficient probabilistic training (p-training) strategy that largely simplifies the training process. The aperture size is set to zero during training. Each training step consists of the following: (1) Randomly choose a scale by predefined probabilities. (2) Scale the receptive circle of each pixel, calculate the composite cones, and divide them into frustums. (3) Calculate each frustum's expected PE and feed them into MLP to predict the density and radiance that represent the frustum. (4) Apply volume rendering on each composite cone to generate pixel colors. (5) Calculate the loss between the rendered colors and the colors of blurred GT. (6) Backpropagate the gradient decent to optimize the MLP parameters.

Although our p-training only applied on special cases (aperture size as zero), experiments show that the trained MLP can well generalize to various cases. At test time, NeRFocus can achieve high-quality defocus effects with adjustable camera pose, focus distance, or aperture size, e.g., by setting the aperture diameter as zero to render large DOF images as NeRF~\cite{nerf} does.

Despite such merits, NeRFocus incurs almost no additional cost on computation or parameters, neither sacrificing performance in rendering large DOF images.

In short, our contributions include:
\begin{itemize}
\item 
NeRFocus, an novel NeRF-based framework that implements thin lens imaging, by which we can render controllable 3D defocus effects.
\item P-training, a probabilistic training strategy that eliminates the requirement of various depth-of-field datasets and vastly simplifies the training process.
\end{itemize}

\section{Related Work}

\subsection{Neural Radiance Field}
Neural Radiance Field (NeRF)~\cite{nerf} is one of the exciting pieces of work that has emerged in recent years at the intersection of neural networks and computer graphics. Previously, researchers have explored the potential of MLP to represent graphic properties implicitly~\cite{curless1996volumetric}, \cite{ren2013global}, \cite{sitzmann2019scene}, \cite{henzler2020learning}. Volumetric representations~\cite{lombardi2019neural}, \cite{niemeyer2020differentiable} and volume rendering~\cite{kajiya1984ray}, \cite{max1995optical} also demonstrated the power of rendering high-quality novel views. 
Further, Mildenhall et al. proposed NeRF~\cite{nerf}, which uses an MLP to represent a continuous volumetric scene function. By explicitly designing the view-dependent effects and using positional encoding, NeRF significantly improved the performance for rendering realistic complex scenes.

Methods has been proposed to further extend the performances of NeRF, e.g., accelerate training or inference time~\cite{neff2021donerf}, \cite{tancik2021learned}, \cite{hedman2021baking}, \cite{wang2021learning}, \cite{lombardi2021mixture}, improve the generalization performance~\cite{yu2021pixelnerf}, \cite{trevithick2021grf}, \cite{wang2021ibrnet}, \cite{chen2021mvsnerf}, enable dynamic scene rendering~\cite{pumarola2021d}, \cite{li2021neural}, \cite{xian2021space}, use unstructured images for training~\cite{martin2021nerf}, \cite{park2021nerfies}, extend NeRF for multiscale and unbounded secens~\cite{neff2021donerf}, \cite{mipnerf}, \cite{barron2021mip}, scene disentanglement and control~\cite{zhang2021editable}, \cite{ost2021neural}, \cite{niemeyer2021giraffe}, \cite{park2021hypernerf}, etc. This paper proposes the first framework that extends NeRF to approximate thin lens imaging, so the defocus effects can be rendered. 

Our method is directly inspired by mip-NeRF~\cite{mipnerf}, a multiscale representation for anti-aliasing rendering. Mip-NeRF extends the pixel's receptive field to a circle. Instead of sampling rays, mip-NeRF sampling conical frustums and use the Integrated Positional Encoding (IPE) to make the encoded frequency components reflect not only the changes of position but also the variance of frustum size. However, mip-NeRF's modeling is based on pinhole imaging, thus can not render lens effects. We explicitly model the thin lens imaging and derive composite cones that can be used to render each pixel into equivalent lens effects.

\subsection{Synthetic Defocus Effects}
In general, high-quality defocus effects heavily rely on expensive wide-aperture lenses. Thanks to advances in machine learning, some approaches have emerged to synthesize defocus effects from single or multiple large DOF images that mobile cameras can easily obtain. A simple solution for synthetic defocus is using a segmentation network~\cite{he2017mask}, \cite{cheng2021masked} to separate the foreground and background, then applying blur operation on the segmented background~\cite{shen2016automatic}, \cite{shen2016deep}, \cite{wadhwa2018synthetic}. 
More general approaches are based on multiplane representations. For instance, one can apply depth estimation~\cite{liu2015learning}, \cite{ranftl2019towards} to divide the scene into several layers of different depths, then use depth-dependent kernels to blur these layers and composite them as an image to synthesize defocus effects~\cite{narain2015optimal}, \cite{mpi}. Similar methods can be applied to dual-lens images where the predicted disparity plays the role of depth~\cite{barron2015fast}, \cite{wadhwa2018synthetic}. These image-based methods simplify the defocus effects in lens imaging as the ``gather" operation on discretized depth layers, i.e., each pixel gathers influences from nearby pixels to form its color, which is usually implemented by convolution with a circular blur kernel. ``Gather" operation may suffer edge artifacts if the depth or disparity estimation is inaccurate. Besides, the multiplane representation is hard to handle the occluded area between its discretized depth layers, making it difficult to synthesize free-view 3D defocus effects.

Recently, Mildenhall et al.~\cite{rawnerf} proposed RawNeRF and integrated it with the multiplane representation to synthesize 3D defocus effects. By training on raw images, RawNeRF can render images with a high dynamic range (HDR) that can be further retouched, such as changing exposure, tone mapping, and focus. To synthesize defocus effects, they first precompute a multiplane representation using the trained RawNeRF, then use the mentioned layer-wise blur operation to synthesize 3D defocus effects. However, NeRF-based precomputation of multiplane representation is either time-consuming or memory-consuming. Instead, NeRFocus, based on precise optical modeling, can directly render effects without post-process. Besides, our method should be compatible with RawNeRF's ``HDR training", which may help improve our visual quality of defocus effects, especially the saliency of defocused bright highlights.

\section{Method}
Although NeRF and its variants achieve excellent performance in novel view synthesis, they can not render defocus effects efficiently. As mentioned earlier in the Introduction, the major difficulties for NeRF to render defocus effects constitute two: (1) The inherent limitations of the pinhole imaging model. (2) The prediction for different sizes of frustum's radiance and density.

Before we articulate how we solve these difficulties, let's briefly review the theory basics in NeRF and thin lens imaging.

\subsection{Rendering in Neural Radiance Field}

The neural radiance field (NeRF)~\cite{nerf} is a field of radiance (i.e., view-dependent color) and density that is consistently represented by a multilayer perceptron (MLP). To calculate the color of a pixel on the sensor plane, NeRF projects a ray $\textbf{r}$ from the pixel to the pinhole and passes it through the scene space, then samples the ray by interval $\delta_{i}$, $i \in \{1,..., N\}$ to form $N$ sample points $\textbf{x}_{i}$. Positional Encoding (PE) is applied on $\textbf{r}$ and each sample point $\textbf{x}_{i}$:
\begin{equation}
\begin{split}
    \gamma(\textbf{x}_{i}) =  \left [ sin(\textbf{x}_{i}),cos(\textbf{x}_{i}),...,sin(2^{L-1}\textbf{x}_{i}),cos(2^{L-1}\textbf{x}_{i}) \right ]^{\top},\\
    \gamma(\textbf{r}) =  \left [
    sin(\textbf{r}),cos(\textbf{r}),...,sin(2^{M-1}\textbf{r}),cos(2^{M-1}\textbf{r}) \right ]^{\top},
    \label{eq:posenc}
\end{split}
\end{equation}
where $\textbf{x}_{i}$ represents the 3D location $(x_{i},y_{i},z_{i})$, $L$ and $M$ are hyperparameters, denoting the number of frequency components in PE. Tancik et al.~\cite{tancik2020fourier} reveal that the MLP is insensitive to high-frequency variance unless explicitly provided, which makes PE essential to the success of NeRF.
The resulting frequency components $\gamma(\textbf{x}_{i})$ and $\gamma(\textbf{r})$ will be fed into the MLP to predict the corresponding color $\textbf{c}_{i}$ and density $\sigma_{i}$:
\begin{equation}
    \left [\sigma_{i}, \textbf{c}_i \right ] = \text{MLP}(\gamma(\textbf{x}_{i}),\gamma(\textbf{r})).
\end{equation}
The predicted colors and densities will be used for discrete volume rendering~\cite{max1995optical} to generate the pixel color $\textbf{C}(\textbf{r})$:
\begin{equation}
\begin{split}
    \textbf{C}(\textbf{r})= \sum_{i=1}^{N}T_{i}(1-\exp(-\sigma_{i}\delta_{i}))\textbf{c}_{i},\\
    where \quad T_{i}=\exp(-\sum_{j=1}\sigma_{j}(\delta_{j}))
\end{split}
\label{eq:rendering}
\end{equation}
The same rendering process is applied on every pixel of the sensor plane and finally composite a complete image. 

Given a sparse set of posed images for training, NeRF chooses a pose for rendering and compares the result with the ground truth (GT) corresponding to the camera pose. Owing to the differentiable volume rendering formulation and the continuity of predicted color $\textbf{c}_{i}$ and density $\sigma_{i}$, the gradient descent of loss function can smoothly backpropagate to the MLP parameters, thus making the MLP easy to converge to a plausible scene representation. In this way, NeRF achieves novel view synthesis without depth supervision.

\subsection{Defocus in Thin Lens Imaging}
\label{subsection:Defocus in thin lens imaging}
\begin{figure*}
  \includegraphics[width=\linewidth]{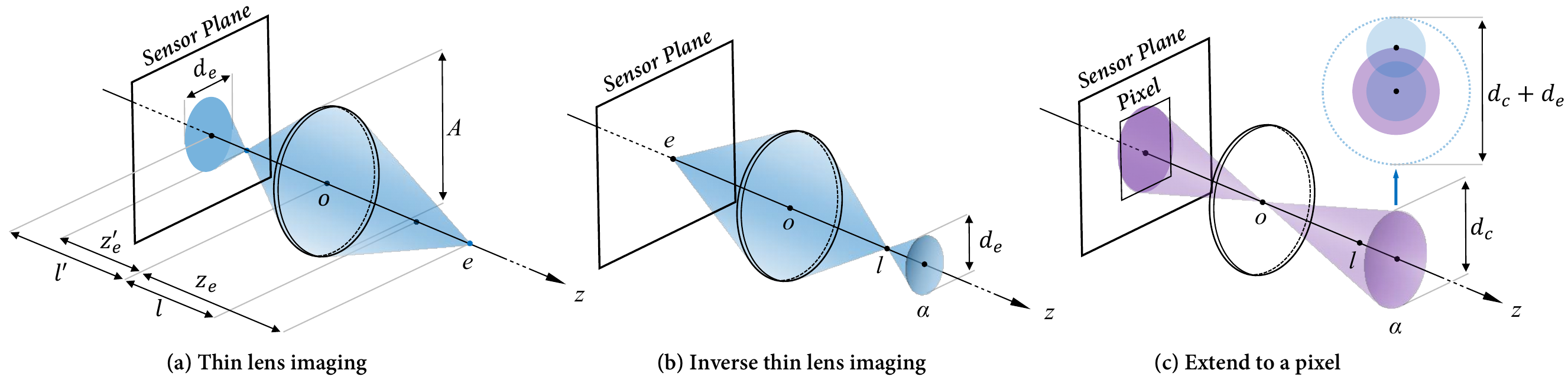}
  \vspace{-5mm}
  \caption{Illustration for modeling thin lens imaging in NeRF. Note all ``distance" in this paper denote the distance along the optical axis, i.e., $z$-axis, by our definition. (a) Defocus in thin lens imaging. If a point is away from the focal plane, its imaging on the sensor plane is a circle rather than a point, called the circle of confusion (CoC). (b) Inverse the thin lens imaging to form a bicone for each point on the sensor plane. The bicone can be seen as the beam path of a point's imaging. Note the ray from the point to the lens center is the axis of the bicone. $\alpha$ denotes the plane ``$z=z_{\alpha}$". $d_{e}$ denotes the bicone's diameter on plane $\alpha$. (c) Use a circle to represent the receptive field of a pixel. The pixel's beam path forms a composite cone, which is overlapped by the beam paths of the points inside the receptive circle. As mentioned earlier, each point's beam path forms a bicone. Note the bicone's axis should cross the lens center, so all the axis of the bicones belonging to this pixel forms a cone (purple region), with $d_{c}$ denoting its diameter on plane $\alpha$. Then we expand the bicones along their axes, so the composite cone's diameter on plane $\alpha$ is exactly the sum of $d_{c}$ and $d_{e}$. Fig.\ref{fig:nerfocus} further illustrates the rendering process of the composite cone.}
  \label{fig:lens imaging}
\end{figure*}

The thin lens model ignores optical effects due to lens thickness and is usually used to simplify the analysis of lens imaging. The part (a) in Fig.\ref{fig:lens imaging} illustrates a thin lens model. Given a thin lens with aperture diameter $A$, focal length $f$, and focus distance $l$, we can calculate the image distance $l'$ following the Gaussian lens equation:
\begin{equation}
    l' = \frac{fl}{f+l}.
\end{equation}
Note that all ``distance" in this paper denotes the distance along the optical axis, i.e., the z-axis by our definition.
The focal plane is at the focus distance $l$ and is orthogonal to the z-axis. The sensor plane is also orthogonal to the z-axis but located at the image distance $l'$. Here we take a point $e$ in the scene space for analysis. If $e$ is away from the focal plane, its imaging on the sensor plane is a circle rather than a point. This circle is called the circle of confusion (CoC) and is the immediate cause for defocus effects. Given $e$ at distance $z_{e}$, we can get its image distance $z_{e}'$ by Gaussian lens equation and use the properties of similar triangles to calculate its CoC diameter $d_{e}$:
\begin{equation}
\begin{split}
    z_{e}' = \frac{fz_{e}}{z_{e}+f} \quad with  \quad \frac{A}{d_{e}} = \frac{z_{e}'}{\left|z_{e}'-l'\right|},\quad
    i.e.,\quad d_{e} = \left|\frac{Af(z_{e}-l)}{z_{e}(f+l)}\right|.
\end{split}
\label{eq:coc size}
\end{equation}
If we move $e$ on the plane ``z=$z_{e}$", its CoC diameter $d_{e}$ is invariant. This attribute is critical for the following derivation.

\begin{figure}[t]
  \centering
  \includegraphics[width=8cm]{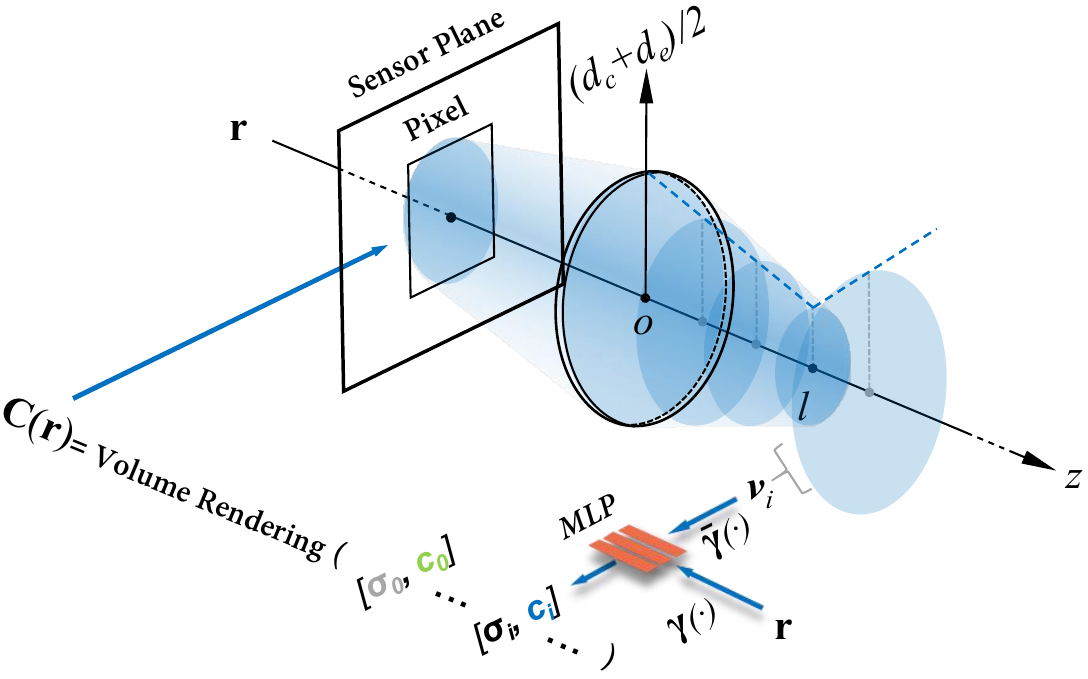}
  \caption{Rendering process of a pixel. The line through the pixel center to the lens center is denoted as $r$, which can uniquely identify the composite cone of the pixel. 
  The blue dotted line denotes the radius. We divide the composite cone into frustums, and use the modified integrated positional encoding (IPE), denoted as $\bar{\gamma}$, to encode each frustum $\textbf{v}_{i}$. 
  Then use an MLP to directly predict the expected radiance and density of each frustum. By volume rendering the predicted results, we can take the expected returned radiance of this composite cone as its corresponding pixel color.}
  \label{fig:nerfocus}
\end{figure}

\subsection{Approximate Thin Lens Imaging in NeRF}
Existing NeRFs are based on a pinhole imaging model, where the radiance of each object point in scene space is directly passed through the pinhole and imaging on the sensor plane as a point, without any lens-like optical refraction. Therefore, the whole scene is in sharp focus and defocus effects can not be rendered.

A typical way to approximate thin lens imaging in ray tracing is to sample enough rays and calculate the average returned radiance. But this will be computational expensive for NeRF, as the rendering of each ray needs dense predictions using MLP.
Instead, we inverse the thin lens imaging to calculate the beam path of each pixel and directly render them. Fig.\ref{fig:lens imaging} and Fig.\ref{fig:nerfocus} illustrate our modeling.

Let's put the point $e$ on the sensor plane for analysis. Following the same assumption in NeRF that the ray path is reversible, we project rays from $e$ to every point on the lens plane. Actually, the corresponding refracted beam path constitutes a bicone that can be described in closed form following the lens equation. Say we have a plane ``z=$z_{\alpha}$" denoted as $\alpha$, the bicone's cross-section on $\alpha$ should be a perfect circle, whose diameter $d_{e}$ can be calculated following Eq.\ref{eq:coc size}, but with different parameters (symmetrically, if take $e$ as the scene point and $\alpha$ to be the sensor plane, this circle can be seen as the CoC of $e$ on the plane $\alpha$):
\begin{equation}
\begin{split}
    d_{e} = \left|\frac{A(l-z_{\alpha})}{l}\right|.
\end{split}
\label{eq:sor size}
\end{equation}
Ideally, we can get the point $e$'s color by calculating the radiance returned by the bicone. But, how to get the color of a pixel? NeRF uses the color of a point to represent the color of a pixel, while in Mip-NeRF~\cite{mipnerf}, Barron et al. extend the receptive field of a pixel as a circle, which achieves better rendering quality. Our method can easily incorporate this extension, as is shown in the (c) part of Fig.\ref{fig:lens imaging}. Say we have a pixel, and the point $e$ is located at the pixel center. We use a circle centered on $e$, with diameter as $d_{0}$, to represent the receptive field of the pixel. Follow the practice in mip-NeRF, we set $d_{0}$ to the width of the pixel scaled by $2/\sqrt{3}$ for all experiments. The pixel's beam path forms a composite cone, which is overlapped by the beam paths of the points inside the receptive circle. 
We project lines from each point inside the circle through the lens center. Each line can be seen as the axis of a bicone. These lines constitute a cone (purple region in Fig.\ref{fig:lens imaging}). Based on parallel similarity, the cone's cross-section on $\alpha$ plane should always be a perfect circle, with diameter:
\begin{equation}
\begin{split}
    d_{c} = \frac{d_{0}z_{\alpha}}{l'} = \frac{d_{0}z_{\alpha}(f+l)}{fl}.
\end{split}
\label{eq:dc}
\end{equation}

Let's take the cross-section on $\alpha$ plane for analysis. Since the cone's cross-section is a circle with diameter $d_{c}$, and each bicone's cross-section is a circle with diameter $d_{e}$, the bicones overlap as a composite cone, whose diameter on $\alpha$ plane is:
\begin{equation}
\begin{split}
    d = d_{c} + d_{e} = \frac{d_{0}z_{\alpha}(f+l)}{fl}+\left|\frac{A(l-z_{\alpha})}{l}\right|,
\end{split}
\label{eq:mip coc size}
\end{equation}
where we can see that four parameters fully characterize the composite cone: pixel's receptive diameter $d_{0}$, aperture diameter $A$, focal length $f$, and focus distance $l$. Then we can get the pixel's color by calculating the composite cone's returned radiance. To simplify the computation, we partition the composite cone by putting the $\alpha$ plane on different distance intervals $z_{i}$, $i \in \{1,..., N+1\}$ to form $N$ conical frustums $\textbf{v}_{i}$. We want to calculate a radiance $\textbf{c}_{i}$ and density $\sigma_{i}$ to represent each frustum $\textbf{v}_{i}$, so we can use Eq.\ref{eq:rendering} to calculate the returned radiance. But the problem is, how should we calculate the representative radiance and density?

In theory, every point inside the frustum may contribute radiance, but their ``importance" may not equal, as the bicones in the composite cone overlap unevenly. Intuitively, a point close to the composite cone axis should contribute more because more bicone is overlapping. Therefore, we model the importance distribution inside frustum $\textbf{v}_{i}$ as a 3D Gaussian distribution $\mathcal{N}(\boldsymbol{\mu}_{i},\boldsymbol{\Sigma}_{i})$, with independent distribution of each coordinate $x, y, z$. 
Theoretically, our frustum's cross-section is parallel to the sensor plane while mip-NeRF's is orthogonal to the cone axis. But these differences bring negligible varies for the encoded value of frustum. So we adopt Mip-NeRF's coordinate system and it's IPE for simplicity. The mean value vector $\boldsymbol{\mu}_{i}$ is composed of $(\mu_{x},\mu_{y},\mu_{z})$, the same value as the frustum's centroid coordinate. The covariance matrix $\boldsymbol{\Sigma}_{i}$ is a diagonal matrix with elements $\sigma_{x},\sigma_{y},\sigma_{z}$, i.e., the variances along each coordinate. These variances are set as the same value as the frustum's uniform distribution variance along each axis.

Therefore, we define the representative radiance $\textbf{c}_{i}$ and density $\sigma_{i}$ of a frustum $\textbf{v}_{i}$ by calculating the expectation of radiance and density of every point inside this frustum:
\begin{equation}
    \left [\sigma_{i}, \textbf{c}_i \right ] = \text{E}_{\mathbf{x}\sim \mathcal{N}(\boldsymbol{\mu}_{i},\boldsymbol{\Sigma}_{i})}[\text{MLP}(\gamma(\mathbf{x}),\gamma(\textbf{r}))].
    \label{eq:emlp}
\end{equation}
A direct way to implement this operation is sampling enough points inside the frustum $\textbf{v}_{i}$ follow the 3D Gaussian, using an MLP to predict their radiance and density, then calculate the average radiance and density.
But this is extremely time-consuming. Inspired by Mip-NeRF, we use an MLP to directly predict the radiance of frustum $\textbf{v}_{i}$, with the expected value of $\gamma(\textbf{x})$ as input. 

Specifically, each dimension $x, y, z$ of the 3D Gaussian is independent of each other. Moreover, each frequency component in positional encoding (PE) is independent of each other as well, thus the distribution of each PE component can be simplified as a 1D Gaussian distribution. Say we have a coordinate x with distribution $\mathcal{N}(\mu_{x},\sigma_{x})$, the expected value of a frequency component $sin(2^{k}x), k\in\{0, ..., L-1\}$, in PE $\gamma(x)$ can be formulated as:
\begin{equation}
    \text{E}_{x\sim \mathcal{N}(\mu_{x},\sigma_{x})}[sin(2^{k}x)]= sin(2^{k}\mu_{x})\exp(-(2^{k}\sigma_{x})^{2}/2),
    \label{eq:ipe analysis}
\end{equation}
which can be easily promoted to each frequency component of each coordinate. In short, the expected value of $\gamma(\textbf{x})$ inside frustum $\textbf{v}_{i}$ can be easily calculated in closed form, denoted as: 
\begin{equation}
    \bar{\gamma}(\textbf{v}_{i}) = \text{E}_{\textbf{x}\sim \mathcal{N}(\boldsymbol{\mu}_{i},\boldsymbol{\Sigma}_{i})}[\gamma(\textbf{x})] .
\end{equation}

Then we use the MLP to predict the radiance and density of frustum $\textbf{v}_{i}$, with $\bar{\gamma}(\textbf{v}_{i})$ and $\gamma(\textbf{r})$ as inputs:
\begin{equation}
    \left [\sigma_{i}, \textbf{c}_i \right ] = \text{MLP}(\bar{\gamma}(\textbf{v}_{i}),\gamma(\textbf{r})).
    \label{eq:mlp predict}
\end{equation}

By volume rendering the predicted colors and densities of all frustums using Eq.\ref{eq:rendering}, we can calculate the radiance returned by a composite cone and use it to represent the corresponding pixel color. Fig.\ref{fig:nerfocus} illustrates the rendering process of a pixel. The above formulation can be extended to every pixel on the sensor plane and finally approximate the optical effects of thin lens imaging.

\subsection{P-Training Strategy}
So far, we have built concise formulations to implement thin lens imaging in NeRF, but here comes another question: how can we make sure that the MLP can correctly predict the color and density for various sizes of frustum? 

A direct solution is to collect enough images taken by a DSLR camera with a zoom lens and apply supervised training~\cite{srinivasan2018aperture}. However, this solution is complex and demands extra datasets. Instead, we generate training datasets based on simple blur operations on the original dataset and apply a probabilistic training strategy based on the multiple blurred datasets. We name our training framework the p-training.

We consider a special case that the aperture is set to zero. Since $d_{e}=0$, $d$ is equal to $d_{c}$, i.e., the composite cone is degrade to a cone. Following Eq.\ref{eq:dc}, $d_{c}$ is proportional to $d_{0}$. If we scale $d_{0}$ using scalar $k$, we are essentially using the scaled composite cone to calculate the expected radiance as the pixel's color. Meanwhile, suppose we use the same scaled Gaussian to calculate the expectation of the pixel's color on the GT image. In that case, this color should be equal to the rendering color of the composite cone if the MLP can predict correctly by our definition in Eq.\ref{eq:mlp predict} and Eq.\ref{eq:emlp}. Thus we obtain a simple equivalence between rendered results and images. Further, if we introduce different scales of $k$ during training, different scales of composite cones will be involved, which constitutes a simple solution to supervise the MLP for correct prediction.

Specifically, given an original dataset $\textbf{D}=\{\textbf{y}_{1}, ..., \textbf{y}_{u}\}$, where $\textbf{y}$ denotes the images inside the dataset, we use different sizes of Gaussian blur kernel $\textbf{k}_{j}$, $j\in \{1, ..., m\}$ to apply convolution operation (denoted as $\ast$) on every image in dataset $\textbf{D}$:
\begin{equation}
    \textbf{D}_{j} = \{\textbf{y}_{1}\ast\textbf{k}_{j}, ..., \textbf{y}_{u}\ast\textbf{k}_{j}\} \quad j \in \{1, ..., m\},
\end{equation}
which forms a set of multi-blur datasets $\{\textbf{D}_{1}, ..., \textbf{D}_{m}\}$.

\begin{figure}[t]
  \centering
  \includegraphics[width=8cm]{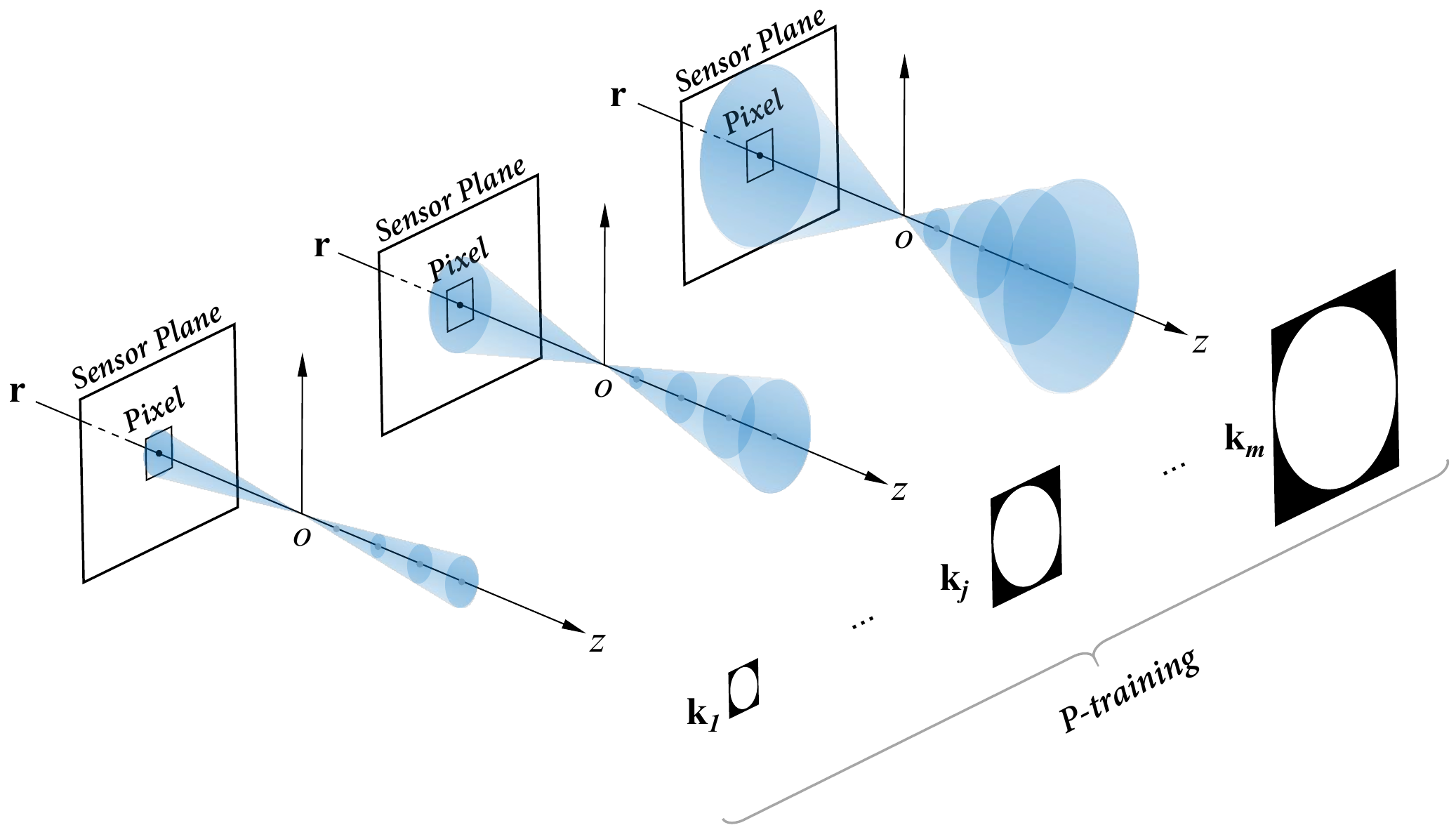}
  \caption{Illustration of our p-training strategy. The aperture diameter is set to zero during training, so $d_{e}$ is equal to zero, the composite cone is degrade as a cone. For each training step, we use predefined probabilities to select a blur kernel $\textbf{k}_{j}$, $j\in \{1, ..., m\}$ randomly. Then, use the diameter of $\textbf{k}_{j}$ to scale up every composite cone's diameters and use $\textbf{k}_{j}$ to blur the original image as the rendering target. This training process will urge the MLP to correctly predict the radiance and density of frustums in different sizes.}
  \label{fig:training}
\end{figure}

As is shown in Fig.\ref{fig:training}, $\textbf{D}_{j} \in \{\textbf{D}_{1}, ..., \textbf{D}_{m}\}$ is randomly chosen with predefined probabilities for each training setp (usually, give datasets with smaller $\textbf{k}_{j}$ the higher chances to be chosen). We use the size of $\textbf{k}_{j}$ to scale the composite cone's diameter. The rendering process is the same as is described before. Following the practice in mip-NeRF\cite{mipnerf}, we use a single MLP to implement a two-step coarse-to-fine training. For each composite cone, we first apply a uniform sampling $\textbf{t}^{c}$ to divide frustums and calculate the ``coarse" rendered color $\textbf{C}(\textbf{r};\Theta, \textbf{t}^{c})$ and the importance distribution along the composite cone's axis, Then apply the importance sampling $\textbf{t}^{f}$ to divide frustums and recalculate the ``fine" rendered color $\textbf{C}(\textbf{r};\Theta, \textbf{t}^{f})$.
The training objective is to minimize the disparity between the blurred GT $\bar{\textbf{C}}(\textbf{r})$ and the composite cone's rendered color:
\begin{equation}
\begin{split}
    \min_{\Theta}\sum_{\textbf{r} \in \mathcal{B}} (\lambda\left \| \bar{\textbf{C}}(\textbf{r})-\textbf{C}(\textbf{r};\Theta, \textbf{t}^{c}) \right \|_{2}^{2}
    + \| \bar{\textbf{C}}(\textbf{r})-\textbf{C}(\textbf{r};\Theta, \textbf{t}^{f}) \|_{2}^{2}),
\end{split}
\end{equation}
where $\Theta$ denotes parameters of the MLP. $\mathcal{B}$ denotes the ray batch. $\bar{\textbf{C}}(\textbf{r})$ is the pixel color in the dataset $\textbf{D}_{j}$, i.e., the blurred pixel color using Gaussian kernel $\textbf{k}_{j}$. $\textbf{t}^{c}$ and $\textbf{t}^{f}$ consist of 128 sampling points, respectively. $\lambda$ is a hyperparameter.

\begin{figure}[h]
  \centering
  \includegraphics[width=6cm]{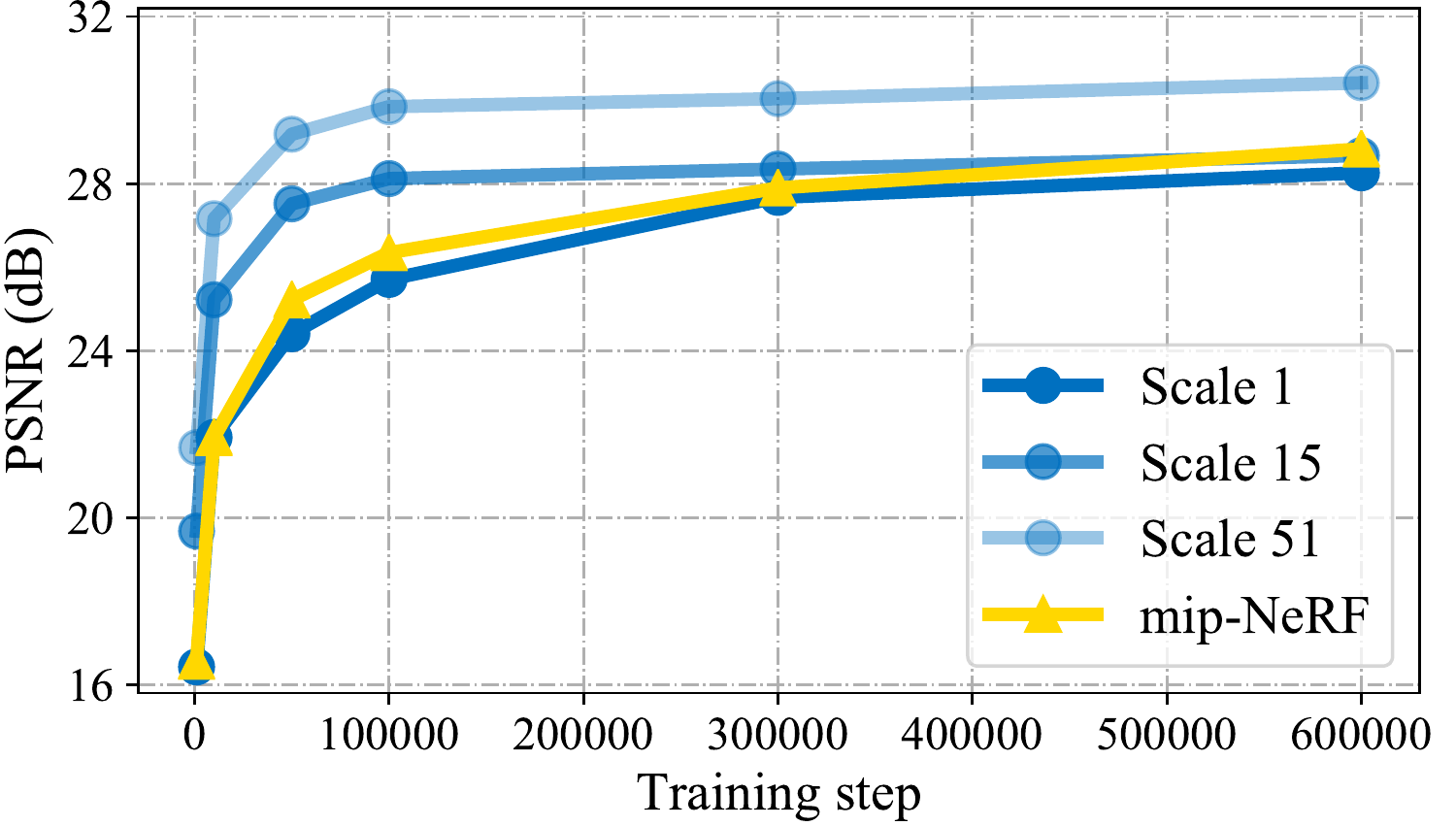}
  \caption{PSNRs by training steps. The yellow line denotes mip-NeRF. The blue lines denote NeRFocus with different scales in p-training. Each scale's PSNR is calculated by the corresponding blurred GT. Note ``scale 1" shares the same GT with mip-NeRF and achieves comparable performance to that of mip-NeRF, which indicates that each scale's learning is not contradictory but complementary.}
  \label{fig:training curve}
\end{figure}

Although our p-training only applied on special cases (aperture diameter as zero and scaled $d_{0}$), experiments show that the trained MLP can generalize to various lens parameters. We guess that this is largely related to the characteristics of IPE. Let's take Eq.\ref{eq:ipe analysis} for example, where $sin(2^{k}\mu_{x})$ can be seen as the original PE. The essence of IPE lies on the coefficient $\exp(-(2^{k}\sigma_{x})^{2}/2)$, which regulates the IPE's value range. If x has a more scattered distribution, i.e., high $\sigma_{x}$, the coefficient $\exp(-(2^{k}\sigma_{x})^{2}/2)$ will approach zero. If the distribution is fixed, a component with a higher frequency $k$ will have a smaller value range. 
Therefore, $\text{E}_{\textbf{x}\sim \mathcal{N}(\boldsymbol{\mu}_{i},\boldsymbol{\Sigma}_{i})}[\cdot]$ can be seen as a low pass filter for $\gamma(\textbf{x})$, considering $\gamma(\textbf{x})$'s different frequency components. Meanwhile, applying $\text{E}_{\textbf{x}\sim \mathcal{N}(\boldsymbol{\mu}_{i},\boldsymbol{\Sigma}_{i})}[\cdot]$ on an image or on scene space will smooth the colors or $\left[\sigma_{i}, \textbf{c}_i \right ]$ pairs. Hence, the training for MLP to satisfy constraint $\text{MLP}(\text{E}[\gamma(\textbf{x})],\gamma(\textbf{r})) = \text{E}[\text{MLP}(\gamma(\textbf{x}),\gamma(\textbf{r}))]$ essentially lets the MLP learn such mapping: let the low-frequency components in IPE map to the blurred color, and gradually approximate the original color as the high-frequency components emerge. This type of mapping is easy for MLP to learn, as is proved in Fig.\ref{fig:training curve}.

Consequently, we can use the trained MLP to correctly predict colors and densities given frustums with different sizes and positions as inputs. By which we can render variant defocus effects. For instance, change the focus distance $l$ to move the DOF forward or backward, change the aperture diameter $A$ to raise or reduce the level of defocus effects. 

\begin{figure*}[h]
  \includegraphics[width=\linewidth]{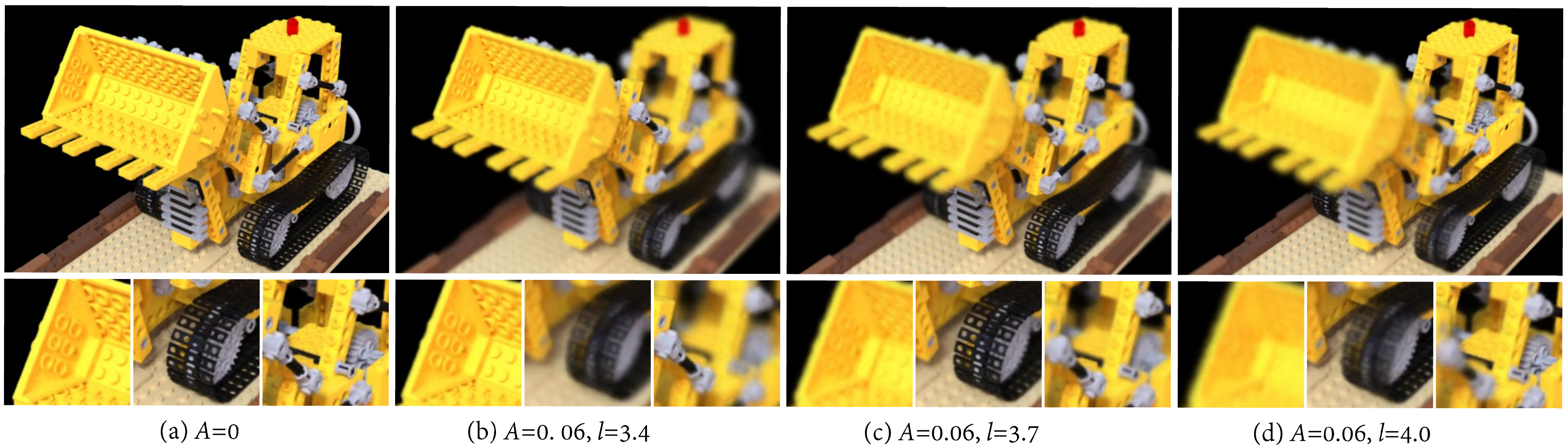}
  \caption{NeRFocus on the lego dataset. The camera pose is fixed for better comparison. We set the aperture diameter $A$ as zero to synthesize image (a) where the whole scene is in sharp focus. We set $A$=0.06, $l$=3.4 to synthesize image (b) where the shovel is in sharp focus. We increase $l$ to 3.7 to move the focus to the caterpillar band, as is shown in (c). We further increase $l$ to 3.7 to move the focus to the gear wheel, as (d) shows. The results imply that our optical modeling has similar properties as a physical lens system.}
  \label{fig:exp1}
\end{figure*}

\begin{figure}[h]
  \centering
  \includegraphics[width=10cm]{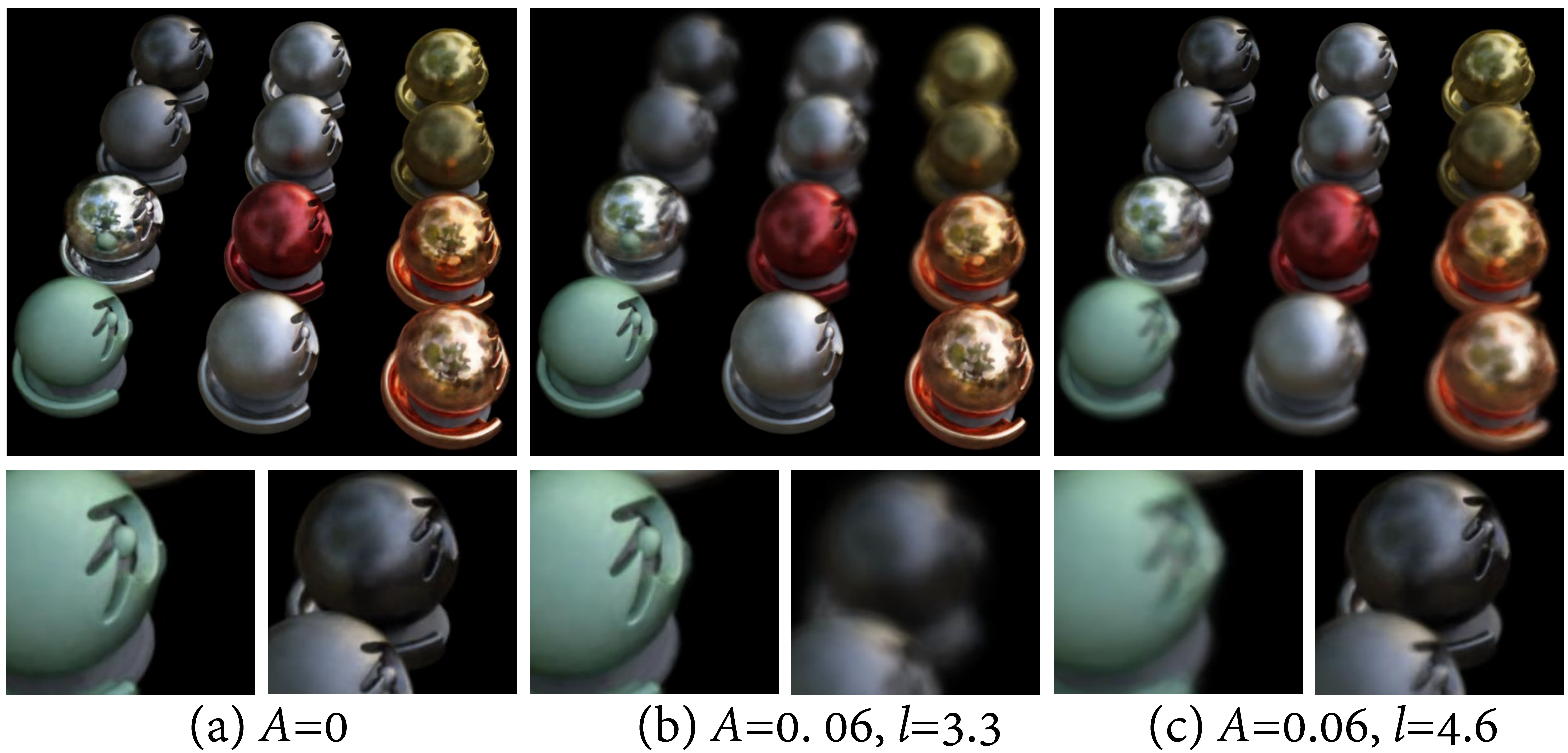}
  \caption{NeRFocus on the materials dataset. We set $A$ as zero to synthesize image (a) where the whole scene is in sharp focus. We set $A$=0.06, $l$=3.3 to synthesize image (b) where the first row is in sharp focus. We increase $l$ to 4.6 to move the focus to the last row, as (c) shows.}
  \label{fig:exp2}
\end{figure}

\begin{figure}[h]
  \includegraphics[width=\linewidth]{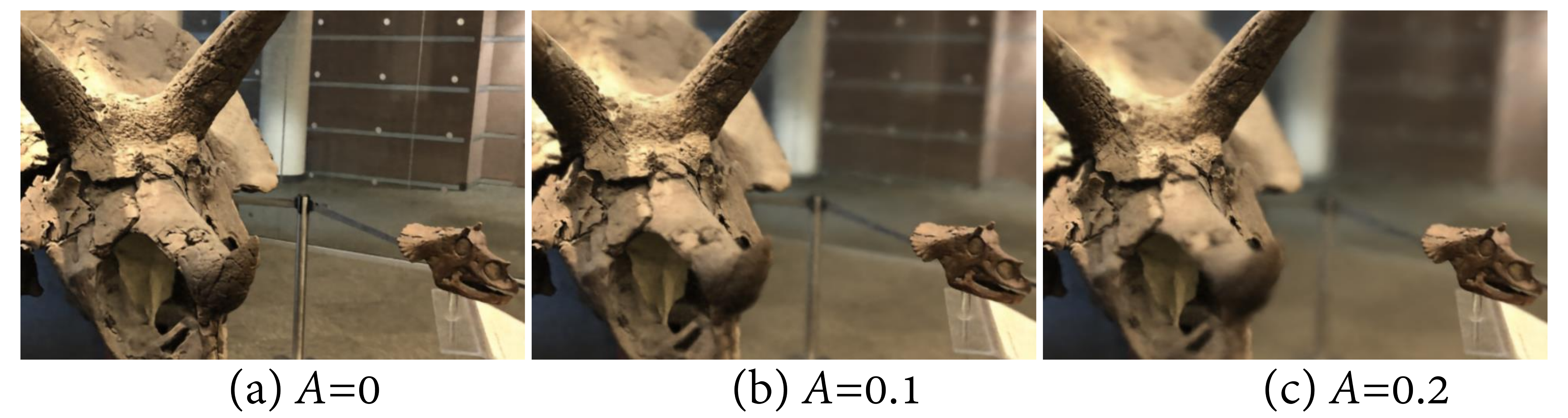}
  \caption{NeRFocus on the horns dataset. We fix the focus distance $l$ at 0.68, and set $A$ as 0, 0.1, 0.2 to synthesize images respectively. (a), (b), (c) shows the cropped results. We can see that the defocus effects get obvious when $A$ increases, while the sharpness of the focused areas are not affected by larger aperture size. Notice that the blur varying along depth shows good continuity.}
  \label{fig:exp3}
\end{figure}

\section{Experiments}
We evaluate NeRFocus on two types of dataset: the synthetic lego and materials dataset presented in NeRF~\cite{nerf} and the camera-captured horns and flower dataset presented in LLFF~\cite{mildenhall2019local}. We set hyperparameters $m$=6, $L$=16, $M$=4 for all experiments. The scales of the blur kernels $\textbf{k}_{j} \in \{1, ..., 6\}$ are set as $\{1,3,7,15,31,51\}$ in units of pixel length, with probabilities $\{0.3,0.2,0.2,0.1,0.1,0.1\}$ respectively. Note we use the original dataset when choosing kernel size 1. For training, we use the Adam optimizer with a logarithmically annealed learning rate that decays from $5\times10^{-4}$ to $5\times10^{-5}$. The batch size of the composite cones is set as 4096 with 600000 training steps (higher training steps may elevate the PSNR but undermine the generalization performance in rendering defocus effects) on a single NVIDIA V100 GPU. For rendering during test time, we fix focal length $f$ as 0.1, so the controllable parameters are aperture diameter $A$, focus distance $l$, and camera pose. This provides simple control for rendering various 3D defocus effects. Theoretically, changing $l$ will result in changing $l'$, which means changing the relative distance from the sensor plane to the lens center, so that all the composite cone's directions have to be recalculated, causing inconvenience for implementation. But since the defocus effects are independent to $d_{c}$, we simply set $l'$ as constant to ignore the change of $l'$ caused by the change of $l$.

\subsection{Quantitative Evaluation}
Fig.\ref{fig:training curve} visualizes the training process on the horns dataset (images are resized to 1008$\times$756). 
Table~\ref{tab:qualitative comparison} shows quantitative comparisons.

Though our approach can render various defocus effects, it incurs no additional costs on computation or parameters. Moreover, by simply setting the aperture $A$ as zero, NeRFocus can render large DOF images as normal NeRF can do, with surprisingly close PSNR.

\begin{table}[t]
    \centering
        \caption{Quantitative comparisons on training time, model parameters, and PSNR. For NeRFocus, we set $A$=0 to calculate the PSNR by the original GT (large DOF). Though NeRFocus achieves defocus effects that mip-NeRF and NeRF cannot do, it neither incurs additional costs on computation and parameters nor sacrifices performance in rendering large DOF images.}
    \begin{tabular}{lccc}
        \hline
            \rule{0pt}{10pt}{\textbf{Method}} & Time(hours)↓ & Params↓ & PSNR↑ \\
        \hline
            \rule{0pt}{10pt}{NeRF (ECCV'20)} & 64.8 & 1192k & 29.00\\
            \rule{0pt}{10pt}{mip-NeRF (ICCV'21)} & 55.6 & 613k & 28.83\\
            \rule{0pt}{10pt}{NeRFocus (ours)} &55.6&613k & 28.26\\
         \hline
    \end{tabular}
    \label{tab:qualitative comparison}
\end{table}

\subsection{Qualitative Evaluation}
Here we demonstrate various defocus effects that NeRFocus can render at test time. As is well known in photography, increasing the aperture diameter will narrow the DOF (i.e., enhance the defocus effects), while increasing the focus distance will move the DOF forward and vice versa. To verify that NeRFocus's rendering also conforms to these optical properties, we fix the camera pose and control the aperture diameter $A$ and focus distance $l$ respectively, to better compare the effects caused by $A$ or $l$. Fig.\ref{fig:exp1} and Fig.\ref{fig:exp2} demonstrate the effects that varying focus distance $l$ brings. Fig.\ref{fig:exp3} shows the effects of changing aperture size $A$. We can see that NeRFocus is in good accordance with the mentioned optical properties, which indicates the effectiveness of our optical modeling. Besides, the blur varying along depth shows good continuity, which implies the excellent generalization performance of the MLP trained by our p-training strategy, especially considering that our p-training only uses five different blur kernels.

\section{Discussion}

We have presented NeRFocus, a framework that implements thin lens imaging in NeRF, by which we can achieve controllable 3D defocus effects. NeRFocus does not need extra datasets with defocus effects. It only requires simple blur operations to generate its training dataset on the original large DOF dataset (e.g., videos taken by drones or cell phones). NeRFocus inherits NeRF's performance in rendering large DOF images and supports rendering shallow DOF images with flexible control on lens parameters. Though with such merits, the training and inference of NeRFocus do not bring extra consumption on time or parameters compared to NeRF or mip-NeRF. Our framework can be seen as a general extension of NeRF or mip-NeRF. e.g., set the aperture size $A$=zero, $(f+l)/fl$ as constant, and use a single kernel size $\{1\}$ for p-training, NeRFocus will degrade to mip-NeRF. We believe NeRFocus has great potential for broader applications and further improvements.

\clearpage
%
%
\bibliographystyle{splncs04}
\bibliography{egbib}
\end{document}